\pdfoutput=1

\documentclass[11pt]{article}

\usepackage[preprint]{acl}

\usepackage{times}
\usepackage{latexsym}

\usepackage[T1]{fontenc}

\usepackage[utf8]{inputenc}

\usepackage{microtype}

\usepackage{inconsolata}

\usepackage{graphicx}

\usepackage{rotating}
\usepackage{adjustbox}
\usepackage{amsmath}
\usepackage{amssymb}

\usepackage{tikz}
\usepackage{pgfplots}
\pgfplotsset{compat=1.18}
\usepgfplotslibrary{colorbrewer}
\usepackage{relsize}
\usepackage{cleveref}
\usepackage{booktabs}
\usepackage[xcolor]{colortbl}
\usepackage{subcaption}
\usepackage{euflag}

\usepackage[round-mode=places, round-precision=2, table-format=2.2]{siunitx}

\title{Pre-trained Language Models Learn Remarkably Accurate \\Representations of Numbers}

\newcommand{\fimuni}{\clubsuit}%
\newcommand{\corecontr}{\dagger}%
\newcommand{\helsinki}{\heartsuit}%
\newcommand{\kinit}{\diamondsuit}%

\author{%
Marek Kadlčík$^{\fimuni\corecontr}$\ \ \
Michal Štefánik$^{\helsinki\fimuni*\corecontr}$\ \ \
Timothee Mickus$^{\helsinki\corecontr}$\ \ \\
\textbf{Michal Spiegel}$^{\fimuni\kinit}$\ \ \
\textbf{Josef Kuchař}$^\fimuni$ \vspace{10pt} \\
    $^\fimuni$TransformersClub @ Faculty of Informatics, Masaryk University \vspace{2pt} \\
    $^\helsinki$Language Technology, University of Helsinki \vspace{2pt}  \\
    $^\kinit$Kempelen Institute of Intelligent Technologies \vspace{2pt} \\
    $^\corecontr${\normalsize Core contributors}
}%

\begin{document}
\maketitle

\def\thefootnote{*}\footnotetext{From October 2025 at National Institute of Informatics Research and
Development Center for Large Language Models, Tokyo, Japan}\def\thefootnote{\arabic{footnote}}

\begin{abstract}

Pretrained language models (LMs) are prone to arithmetic errors. Existing work showed limited success in probing numeric values from models' representations, indicating that these errors can be attributed to the inherent unreliability of distributionally learned embeddings in representing exact quantities.
However, we observe that previous probing methods are inadequate for the emergent structure of learned number embeddings with sinusoidal patterns. 

In response, we propose a novel probing technique that decodes numeric values from input embeddings with \textit{near-perfect accuracy} across a range of open-source LMs. This proves that after the sole pre-training, LMs represent numbers with remarkable precision. 
Finally, we find that the embeddings' precision, judged by our probe's accuracy, explains a large portion of LM's errors in elementary arithmetic, and show that aligning the embeddings with the pattern our probes discover can mitigate these errors.

\end{abstract}

\section{Introduction}

The landmark paper of \citet{brown2020languagemodelsfewshotlearners} showed that generic neural networks trained on text prediction alone could develop surprising arithmetic capabilities.
In the years since, this observation has flourished into a large and vibrant field interested in the arithmetic reasoning capabilities of Transformers \citep{ahn2024largelanguagemodelsmathematical},
rife with research opportunities ranging from interpretability work \citep{10675800} to solving Olympiad-level problems in mathematics \citep{li2025provingolympiadinequalitiessynergizing}.
Yet this work has also underscored the limitations of LMs on arithmetic tasks:
Previous studies have explored how models can benefit from incorporating precise numeric representations \citep{feng2024numericalprecisionaffectsmathematical}, or offloading the arithmetic computation to a tool \citep{toolformer,calcformers}, suggesting that their native learned representations are not reliable. 
Other works \citep{kantamneni2025languagemodelsusetrigonometry,llm-fourier-addition} have inspected such learned representations directly and tried to understand how models use them. Although model probing methods showed some success in interpreting numeric values from model representations~\citep{zhu-etal-2025-lm-num-linears}, the accuracy of those methods is low, suggesting that learned representations are highly imprecise.

In this paper, we push back on this interpretation: we show that a probe with the \textit{right kind of inductive bias} can retrieve numeric information from number embeddings with \textit{near-perfect accuracy} across an extensive range of LMs, spanning the Llama~3 \cite{llama3}, Phi~4 \cite{phi4} and OLMo~2 \cite{olmo2} series and ranging from 1B to 72B parameters.
Given that number embeddings usually follow a sinusoidal wave-like pattern \citep{nanda2023progress, kantamneni2025languagemodelsusetrigonometry}, this characteristic must be accounted for when designing probes.

We further show how these insights can be leveraged to improve performances on arithmetic reasoning: errors on addition and subtraction tasks can often be matched with an inability of the probe to retrieve the expected numerical information for a given embedding, and demonstrate that intervening on number embeddings such that they more cohesively follow the pattern of other number embeddings can directly improve arithmetic performances.
Lastly, we document edge cases that do not fall within this previously understood pattern: in particular, OLMo2 32B \citep{olmo2} learns embeddings that are not sinusoidal-like, despite a high success rate on arithmetic tasks.

\section{Related Work}

One line of work focuses on incorporating numerical values directly into token representations, providing LMs with a prior. \citet{charton2022lingebratransformers} explores different number encodings based on scientific notation for training LM solvers of linear algebra problems. \citet{golkar2023xval} propose representing numbers as a learned <NUM> token scaled by the number scalar value, demonstrating how models can adopt this scheme for regression tasks.

Another line of work investigates how models learn to represent and process numerical information. 
\citet{nanda2023progress} show that a transformer with one-hot encoding trained from scratch on modular addition discovers Fourier basis and its computation is interpretable in trigonometric functions. \citet{kantamneni2025languagemodelsusetrigonometry} discover an analogous circuitry for (non-modular) addition in a general pretrained language model, and find that its intermediate representations combine both linear and periodic components, reminiscent of a helix structure. \citet{llm-fourier-addition} further identifies subcomponents of the addition circuitry implemented by the attention mechanism and feedforward layers. \citet{zhu-etal-2025-lm-num-linears} demonstrate that hidden states of pretrained language models can be approximately decoded with a linear (or multi-layer) probe to estimate the logarithm of the number value. Although the probe outputs correlate with the target value, decoding achieves low accuracy.
Recently, \citet{levy-geva-2025-language} show success in recovering the values of \textit{digits} from internal representations of intermediate layers, hinting on a more generalized, circular pattern in representations of numbers.

In summary, prior works suggest that language models \textit{attempt} to encode numerical information into token representations during pretraining, but their precision is rather limited. However, we hypothesize that this perception stems from inadequate probing methods, and learned representations are much more precise than previously estimated.

\section{Recovering numerical information from number embeddings}
\label{sec:probes}
We study LMs from the Llama~3 \cite{llama3}, Phi 4 \cite{phi4}, and OLMo~2 \cite{olmo2} series, ranging from 1B to 72B parameters.  
Wide selection allows us to verify the validity of our observations across a panel of models sharing the characteristic of representing all integers between 0 and 999 with unique tokens.

\paragraph{Motivations.} The central and foremost point to address is whether the embeddings representing specific numbers in LMs contain the numeric information of the value they represent.
In practice, this is best addressed with a \textit{probing} setup: If embeddings do contain numerical information, we should be able to learn a decoding function from number embedding to the corresponding integer value.
Probing as a methodology comes with its own set of caveats: probes should be kept as simple as possible, and their expressivity should be compared against baseline benchmarks \citep{hewitt-liang-2019-designing}.
Our specific use case adds further constraints: in particular, we have only one instance per LLM of each integer representation, viz., there is only one vector for the token \texttt{42}.
This rules out naive classifier implementations, as we aim for the probe to generalize to entirely unseen classes.

\paragraph{Probe architectures.}
We consider four probes: 
\begin{align}
    f_\mathrm{lin}(\mathbf{x}) &= \mathbf{a}^T\mathbf{x} + b \label{eq:lin}\\
    f_{\log\,\mathrm{lin}}(\mathbf{x}) &= \exp\left(\mathbf{a}^T\mathbf{x} + b\right) - 1\label{eq:loglin}\\
    f_{\sin}(\mathbf{x}) &= (\mathbf{W}_\mathrm{out}\mathbf{S})^T(\mathbf{W}_\mathrm{in}\mathbf{x}) \label{eq:sin}\\
    f_\mathrm{bin}(\mathbf{x}) &= (\mathbf{W}_\mathrm{out}\mathbf{B})^T(\mathbf{W}_\mathrm{in}\mathbf{x}) \label{eq:bin}
\end{align}
\noindent where $\mathbf{a}$, $b$, $\mathbf{W}_\mathrm{in}$, and $\mathbf{W}_\mathrm{out}$ are learned parameters, whereas $\mathbf{S}$ and $\mathbf{B}$ are means of injecting inductive biases in the linear classifiers $f_{\sin}$ and $f_\mathrm{bin}$:
\begin{align*}
    \mathbf{S}_{ij} =& \begin{cases}
        \sin(i e^j 1000 / d) \qquad \mathrm{if~} j \equiv 0 \mod 2 \\
        \cos(i e^{j+1} 1000 / d) \quad \mathrm{if~} j \equiv 1 \mod 2
    \end{cases} \\
    \mathbf{B} =& \begin{bmatrix}
        0 & \dots & 0 &  0 & 1\\
        0 & \dots & 0 &  1 & 0\\
        0 & \dots & 0 &  1 & 1\\
        \vdots
    \end{bmatrix}
\end{align*}
I.e., the $i$\textsuperscript{th} row of $\mathbf{B}$ corresponds to the integer $i$ expressed in binary, whereas $\mathbf{S}$ is defined as a Fourier basis, suggested by~\citeauthor{llm-fourier-addition} as the hidden structure learned by pretrained models. 
The matrices $\mathbf{S}$ and $\mathbf{B}$ thus allows us to partition the label projection of the classifier into three components: a learned projection $\mathbf{W}_\mathrm{in}: \mathbb{R}^d \to \mathbb{R}^h$ to project the number embeddings into a reduced low-dimensional space, a fixed matrix ($\mathbf{S}$ or $\mathbf{B}$) allowing us to encode integers using an \textsl{a priori} scheme, and a learned projection $\mathbf{W}_\mathrm{out}: \mathbb{R}^d \to \mathbb{R}^h$ mapping these \textsl{a priori} representations onto the same space as the reduced embeddings.
Intuitively, $W_{in}$ uncovers the underlying hidden structure of the learned embeddings, while $W_{out}$ expresses it in terms of interpretable a priori basis, which allows us to generalize to unseen tokens.

\paragraph{Implementation.}
We evaluate the probes in \Cref{eq:lin,eq:loglin,eq:sin,eq:bin} using a cross-validation setup with 20 folds.
We report their accuracy measured by rounding the output of the regression probes \Cref{eq:lin,eq:loglin} to the nearest integer, or by retrieving the index of the row in $\mathbf{S}$ or $\mathbf{B}$ that maximizes the output distribution of the classifier probes \Cref{eq:sin,eq:bin}.
We control the validity of our probes by ensuring that they reach an accuracy of 0 for standard Gaussian vectors as well as for a random permutation of the embeddings.
Parameters for regressions are estimated using a least-squares algorithm;
whereas our classifiers' parameters are optimized with Adam with a learning rate of $0.0001$, weight decay of $0.001$, and $\beta = (0.9, 0.999)$. We choose a hidden dimension of 100. The classifiers are optimized to distinguish output \textit{only between training tokens}, and during testing, must choose between all tokens. The probes are optimized until loss converges on a validation split separate from the testing split.

We release an implementation and training recipes for the new probes, including all configurations we use, in the project's GitHub repository.\footnote{\url{https://github.com/prompteus/numllama}}



\begin{figure}[th]
\resizebox{\columnwidth}{!}{
\begin{tikzpicture}
    \tikzstyle{every node}=[font=\footnotesize]
    \begin{axis}[
        ybar, ymin=0, ymax=1.2,
        symbolic x coords={olmo1b, olmo7b, olmo13b, olmo32b, llama1b, llama3b, llama8b, llama70b, phi, phimini},
        bar width=.1cm,
        ytick={0,0.25,0.50,0.75,1.0,1.25},
        yticklabels={0.00,0.25,0.50,0.75,1.00},
        xticklabels={,\bf OLMo 2 1B, \bf OLMo 2 7B, \bf OLMo 2 13B, \bf OLMo 2 32B, \bf Llama 3 1B, \bf Llama 3 3B, \bf Llama 3 8B, \bf Llama 3 70B, \bf Phi 4 15B, \bf Phi 4 4B},
        enlarge y limits={upper=0.75},
        legend style={at={(0.5,1.05)},
            anchor=south,legend columns=-1,
            /tikz/every even column/.append style={column sep=0.125cm}
        },
        nodes near coords always on top/.style={
            scatter/position=absolute,
            positive value/.style={
                at={(axis cs:\pgfkeysvalueof{/data point/x},\pgfkeysvalueof{/data point/y})},
            },
            negative value/.style={
                at={(axis cs:\pgfkeysvalueof{/data point/x},0)},
            },
            every node near coord/.append style={
                check values/.code={%
                    \begingroup
                    \pgfkeys{/pgf/fpu}%
                    \pgfmathparse{\pgfplotspointmeta<0}%
                    \global\let\result=\pgfmathresult
                    \endgroup
                    %
                    %
                    \pgfmathfloatcreate{1}{1.0}{0}%
                    \let\ONE=\pgfmathresult
                    \ifx\result\ONE
                        \pgfkeysalso{/pgfplots/negative value}%
                    \else
                        \pgfkeysalso{/pgfplots/positive value}%
                    \fi
                },
                check values,
                anchor=west,
                rotate=90,
                font=\tiny,
                /pgf/number format/fixed,
                /pgf/number format/zerofill,
                /pgf/number format/precision=2,
                xshift=-0.5ex,
                color=black,
            },
        },
        nodes near coords={
            \pgfmathprintnumber[fixed zerofill,precision=2]{\pgfplotspointmeta}
        },
        nodes near coords always on top,
        height=4cm,
        width=1.5\columnwidth,
        xtick align=inside,
        label style={font=\tiny},
        cycle list/RdYlBu-7,
        every axis plot/.append style={
            fill,
        },
        x tick label style={rotate=90,anchor=east}
    ]

\addplot+ [draw=black,] coordinates {
(olmo1b,0.101772) 
(olmo7b,0.137434) 
(olmo13b,0.12354) 
(olmo32b,0.069213) 
(llama1b,0.327461) 
(llama3b,0.405687) 
(llama8b,0.183672) 
(llama70b,0.327901) 
(phi,0.396758) 
};

 \addplot+ [draw=black,] coordinates {
(olmo1b,  0.593798)     
(olmo7b,  0.963969)     
(olmo13b,  0.978889)     
(olmo32b,  0.069444)     
(llama1b,  0.99629)     
(llama3b,  0.990328)     
(llama8b,  0.956205)     
(llama70b,  0.99902)     
(phi,  0.942125)     
};

 \addplot+ [draw=black,] coordinates {
(olmo1b, 0.009056 )
(olmo7b, 0.025435 )
(olmo13b, 0.019934 )
(olmo32b, 0.003007 )
(llama1b, 0.030979 )
(llama3b, 0.036161 )
(llama8b, 0.009707 )
(llama70b, 0.02118 )
(phi, 0.012287 )
};

 \addplot+ [draw=black,] coordinates {
(olmo1b, 0.023097)
(olmo7b, 0.025562)
(olmo13b, 0.021555)
(olmo32b, 0.004933)
(llama1b, 0.054102)
(llama3b, 0.05488)
(llama8b, 0.025237)
(llama70b, 0.021317)
(phi, 0.019778)
};

\legend{$f_\mathrm{bin}$\qquad\null,$f_{\sin}$\qquad\null,$f_\mathrm{lin}$\qquad\null,$f_{\log\,\mathrm{lin}}$}
    \end{axis}
\end{tikzpicture}
}
\caption{Overview of probes' accuracy ($\uparrow$).}
\label{fig:prob acc}
\end{figure}

\paragraph{Results.} We summarize performances, measured in terms of accuracy, in \Cref{fig:prob acc}.
Crucially, we are almost systematically able to retrieve the integer value corresponding to the embedding's number with very high accuracy.
Another salient observation is that $f_{\sin}$ consistently outperforms all other probe architectures including the regression probe used in previous work of~\citet{zhu-etal-2025-lm-num-linears}, contradicting their finding that LMs learn to encode numbers linearly.
Explaining the success of the Fourier basis, we note that other prior literature has suggested that sinusoidal features are used for arithmetic computation in LMs \citep{llm-fourier-addition}.
Adding onto this, we can also stress that, qualitatively, most of the models' whose number embeddings we survey here exhibit wave-like patterns in a PCA projection and have sparse Fourier transform, confirming regularity in the hidden structure. See \Cref{fig:sinusoidal embs,fig:pca-fourier} in \Cref{sec:appendix} for visualizations of PCA and its Fourier transform. 
Notably, OLMo 2 32B is the only model with low resemblance of the pattern, which is consistent with the low performance of its sinusoidal probe.

\paragraph{Analysis.} To verify that our sin-base probes indeed reach their superior accuracy by learning to extract a generalized, sin-like representation from models' representations, we analyse the encoded representations that trained sin probes produce as the output of $\mathbf{W}_\mathrm{in}$. We experiment with two training settings: (i) using L1 regularization --- encouraging sparsity, and (ii) using L2 regularization --- encouraging the employment of a broader scale of input features. We note that in both of these settings, the probes achieve almost identical generalization capacity as assessed by their accuracy on unseen inputs (embeddings of numbers).

Figure~\ref{fig:probe_weights} in Appendix~\ref{appx:probe_weights} displays the resulting representations for model embeddings of Llama3 1B associated with different numeric values. 
We can observe that the L1- and L2-regularized probes learn a substantially distinct representational pattern. We hypothesize that a main difference between probes trained with different regularizations is that the L1 probe learns to follow a broader scale of distinct frequencies, while the L2 probe follows similar frequencies shifted by a different constant. Nevertheless, in both of the cases, the probe learns a projection into a wave-like pattern across input numbers, thus successfully following their injected inductive bias.

\section{Leveraging numerical information from number embeddings}
\label{sec:add and subtract}

\paragraph{Motivation.} Having established that number embeddings do encode retrieval numerical information about the integers they represent, we now turn to \textbf{how this numerical information is leveraged} by LMs to perform arithmetic tasks.
We study the zero-shot performances of a subset of our models on addition and subtraction tasks.
We define our addition task as taking any pair of integers $x_1$, $x_2$ such that $0 < x_i < 500$ as input, and computing the expected output $x_1 + x_2$.
The subtraction task is defined by taking as inputs any pair $x_1$, $x_2$ such that $0 < x_2 < x_1 < 1000$, and computing the expected output $x_1 - x_2$. 

\paragraph{Performance.}
To perform the arithmetic tasks, we conduct minimal prompt engineering: we systematically evaluate a handful of natural language prompts for their accuracy in a zero-shot setting, and then select the highest-performing for subsequent analyses. Due to computational costs, we ignore the two largest models (OLMo2 32B and Llama 3 70B).
All prompts are listed in \Cref{sec:details}, see \Cref{tab:prompts addition}  for addition and \Cref{tab:prompts subtraction} for subtraction.

\begin{table}[ht]
    \centering
    \resizebox{\columnwidth}{!}{
    \begin{tabular}{>{\bf}l *{7}{S@{{~}}}}
         & {{\rotatebox{90}{\textbf{OLMo2 1B}}}} & {{\rotatebox{90}{\textbf{OLMo2 7B}}}} & {{\rotatebox{90}{\textbf{OLMo2 13B}}}} & 
         {{\rotatebox{90}{\textbf{Llama 3 1B}}}} & {{\rotatebox{90}{\textbf{Llama 3 3B}}}} & {{\rotatebox{90}{\textbf{Llama 3 8B}}}} & 
          {{\rotatebox{90}{\textbf{Phi 4 15B}}}}  
          \\
         \toprule
        Add. & 21.39 & 	1.12&  	0.17& 	2.58& 	0.45& 	0.25 & 0.0 \\
        Sub. & 28.12& 	0.36& 	0.16& 	1.43& 	0.03& 	0.01 & 0.0 \\
        \bottomrule
    \end{tabular}}
    \caption{Overview of error rates (\%, $\downarrow$) on arithmetic tasks in zero-shot setting. }
    \label{tab:arithmetic-perfs}
\end{table}

An overview of the error rates from the LMs we study is listed in \Cref{tab:arithmetic-perfs}. 
As is apparent, most models achieve high degrees of performance (except for OLMO 2 1B); we also observe a trend towards fewer errors for models with more parameters.

\paragraph{Error analysis.}
To assess how numerical information and arithmetic performance are linked, we evaluate whether the errors we see in these arithmetic tasks are associated with defects of the number embeddings used as inputs.

We measure the error rate on the downstream addition and subtraction task in two separate cases -- in the first case, both input tokens 
are decodable by the probe, in the second case, at least one value is not. The results can be seen in~\Cref{tab:arith-perfs-grouped}.

\begin{table}[ht]
    \centering
    \resizebox{\columnwidth}{!}{
    \begin{tabular}{>{\bf}l *{7}{S@{{~}}}}
        &
         {{\rotatebox{90}{\textbf{OLMo2 1B}}}} & {{\rotatebox{90}{\textbf{OLMo2 7B}}}} & {{\rotatebox{90}{\textbf{OLMo2 13B}}}} & 
         {{\rotatebox{90}{\textbf{Llama 3 1B}}}} & {{\rotatebox{90}{\textbf{Llama 3 3B}}}} & {{\rotatebox{90}{\textbf{Llama 3 8B}}}} & 
          \\
         \toprule
        \multicolumn{1}{l}{\textit{Addition}} \\
        \; decodable    & 20.30 & 	0.98 & 	0.16 & 	2.48 & 	0.46 & 	0.24       \\
        \; undecodable  & 23.33& 	1.84& 	0.20& 	14.86& 	0.28& 	0.28   \\
 	    \midrule
        \multicolumn{1}{l}{\textit{Subtraction}} \\
        \; decodable    & 24.61& 	0.32& 	0.19& 	1.43& 	0.03& 	0.01 \\
        \; undecodable  & 31.45 & 0.53 & 0.13& 0.90 & 0.04 & 0.0   \\
        \bottomrule
    \end{tabular}}
    \caption{Downstream arithmetic error rate (\%, $\downarrow$) given that (1) all tokens are decodable, and (2) at least one token is non-decodable. Results are measured on all possible input combinations. Phi is omitted because it does not make errors.}
    \label{tab:arith-perfs-grouped}
\end{table}

The results show that models tend to make more errors when the input embeddings are misaligned with the pattern used by the probe, as undecodable inputs lead to higher error rates in 8 out of 12 configurations. The effect is more prominent for models with substantial error rates, such as OLMo2 1B.


\paragraph{Direct intervention.}
We hypothesize that embeddings of tokens that our probes can not correctly decode diverge from the model's robust representation scheme and thus contribute to errors in arithmetic tasks. With this motivation, we test whether a direct intervention on the embeddings of these tokens
can improve models' performance on arithmetic. In practice, we start from the $f_{\sin}$ probes described in \Cref{eq:sin} and trained for Llama 3 1B and freeze all probe parameters. We then perform gradient descent to optimize the \textit{embeddings} of all incorrectly decoded tokens (namely \texttt{0}, \texttt{4}, \texttt{977} and \texttt{999}) with respect to the probe decoding loss, aiming to align those tokens with the overall pattern discovered by the probe.

We finally measure how this embedding intervention impacts model error rate on addition and multiplication tasks involving these four tokens as one of the inputs or expected outputs, using the model's best-performing template (a set of our experimental templates is listed in Table~\ref{tab:prompts addition}).

We find that in additions involving these assumably divergent tokens, our intervention reduces 26\% of errors (from 17.6\% to 13.0\%). In multiplications, our intervention brings error reduction by 9.4\% (from 8.5\% to 7.7\%).
This experiment, while of an anecdotal scale determined by a low error rate of our probes, shows that more accurate probes of models' representations can also guide direct refinements of models' possibly imprecise embeddings, aligning them with the model's general hidden structure and bringing improvements in accuracy of the model's predictions.

\section{Conclusion}
\label{sec:conclusion}

In this paper, we have inspected the embedding representations for number tokens across a range of widely used open-source LMs. 
Our observations consolidate a growing body of studies showcasing how LMs learn sinusoidal hidden structure in number representations. Building upon this observation, we design a probing method leveraging this structure that decodes LMs' embeddings with \textit{near-perfect} accuracy across multiple models, demonstrating that the quality of numeric representations in pretrained LMs was strongly underestimated in previous work.
Still, we find a model (OLMo 2 32B) that deviates from this pattern, calling into question the generalizability of the conclusions of works such as \citeposs{llm-fourier-addition}.
Finally, 
we show that the preciseness of embeddings relative to the sinusoidal pattern can explain a proportion of practical errors on arithmetic tasks, especially when models fail to align closely with this sinusoidal pattern. 

Furthermore, we demonstrate improved accuracy on those tasks by aligning imprecise embeddings to the model's learned embedding pattern. To some extent, our findings curtail the validity of offloading approaches for numerical reasoning \citep{toolformer,calcformers}: showing that their initial premise --- of models not learn accurately representations of numbers --- is incorrect. 
We hope that our findings will motivate future work to rigorously compare relative advantages of tool-using models in terms of computational efficiency, and challenge future work towards the \textit{data} \cite{stefanik-etal-2024-concept} and \textit{architecture} refinements \cite{spiegel2025attendperishbenchmarkingattention} accelerating more efficient learning of accurate representations of exact elements of language.

\section*{Acknowledgements}
This work is supported by the Research Council of Finland through project No.~353164 ``Green NLP -- controlling the carbon footprint in sustainable language technology''.
\euflag \ This project has received funding from the European Union’s (EU) Horizon Europe research and innovation programme under Grant agreement No. 101070350 and from UK Research and Innovation
(UKRI) under the UK government’s Horizon Europe funding guarantee (grant number 10052546). The contents of this publication are the sole responsibility of its authors and do not necessarily reflect the opinion of the EU.

\section*{Limitations}

Our work, while demonstrating the remarkable accuracy of number embeddings in pre-trained language models, comes with several limitations that warrant consideration for future research.

First, our probing method, though highly effective for many models, relies on an assumed hidden structure of models' learned representations, and therefore expects a broad a priori understanding of models' representation space. This necessarily limits the applicability of our approach to models where a known structure exists; Our results aim to show that some language models indeed \textit{do} exhibit alternativel encoding schemes, exemplified by OLMo 2 32B that, ableit being highly accurate in arithmetics, can not be accurately probed by our sinusoidal probes.

Second, our intervention method was performed on a small-scale experiment, and its generalization across a large suite of models remains an object for future work.

Third, even when we do not perform any pretraining of models, reproducing our experiments requires access to computational resources. We estimate that replicating all our results requires around several hundred GPU hours.

Fourth, our analysis targets model embeddings. It is thus limited to single-token representations, and does not address the inner mechanisms of numeric information processing in large language model. This area also calls for further research.

While we recognize the ethical risks associated with AI research, given that our paper focuses on fundamentals of internal representations of numbers within pre-trained language models and their immediate impact on basic arithmetic tasks, broader societal ethical concerns like bias, discrimination, privacy, or job displacement are not directly relevant. Our research operates at a fundamental level of understanding how models encode numerical information, rather than exploring their application or misuse in real-world systems with downstream societal consequences.

\bibliography{acl_latex}

\appendix

\section{Supplementary visualization}

\subsection{Wave-like patterns in embeddings}
\label{sec:appendix}

\begin{figure*}[th]
    \centering
    \includegraphics[width=1.0\linewidth]{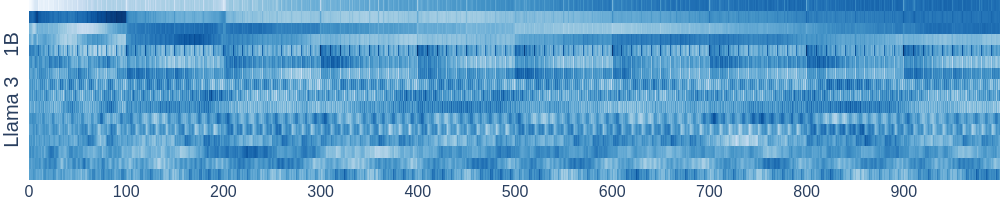}
    \includegraphics[width=1.0\linewidth]{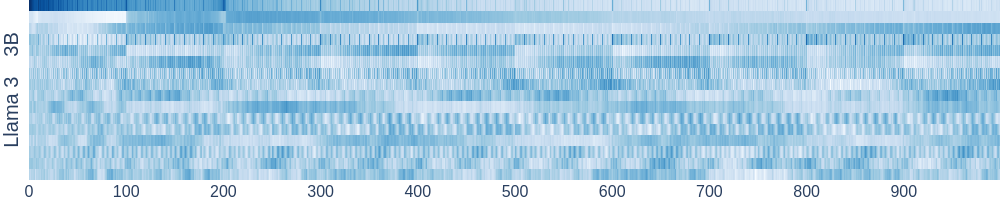}
    \includegraphics[width=1.0\linewidth]{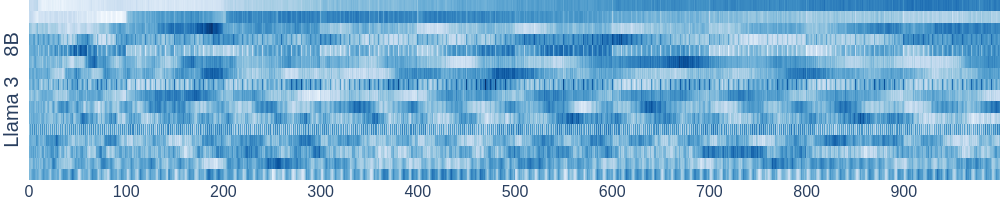}
    \includegraphics[width=1.0\linewidth]{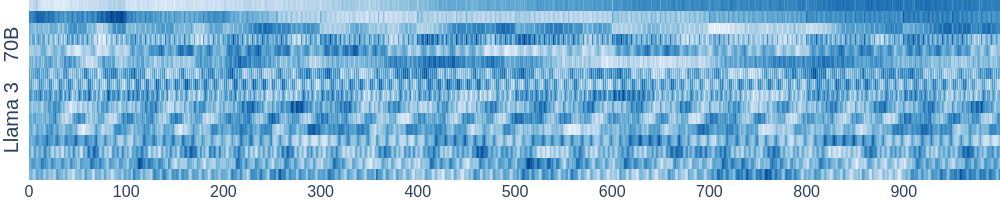}
    \includegraphics[width=1.0\linewidth]{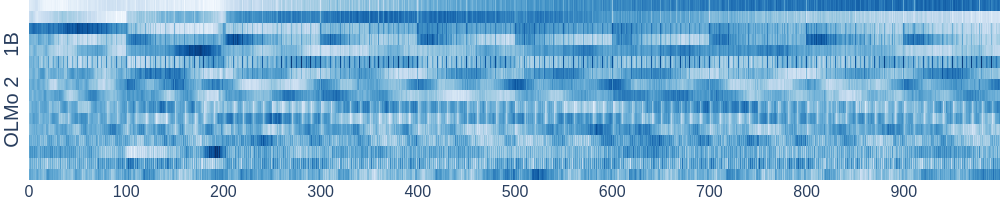}
    \includegraphics[width=1.0\linewidth]{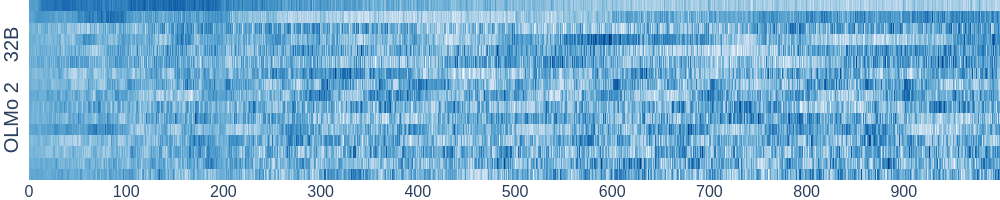}
    \includegraphics[width=1.0\linewidth]{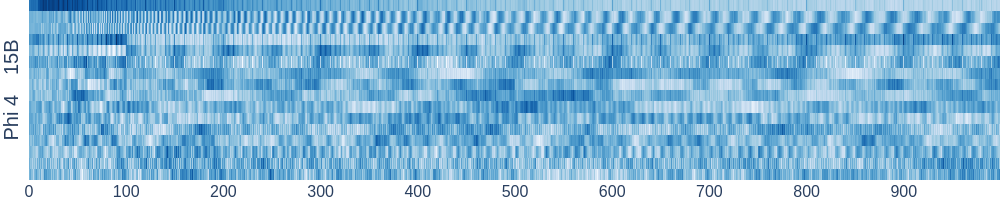}
    {\small Token Value}
    \caption{Visualization of PCA (DIM=16) reduced number embeddings, selected models. Although most model exhibit relatively regular wave-like patterns, OLMO 2 32B exhibit little regularity.}
    \label{fig:sinusoidal embs}
\end{figure*}
\Cref{fig:sinusoidal embs} displays the sinusoidal patterns in Llama 3 70B and OLMo2 13B after PCA dimensionality reduction. For clarity, we only include the first 16 principal components.

\begin{figure*}
    \begin{subfigure}[h]{0.5\linewidth}
        \includegraphics[width=1.0\linewidth]{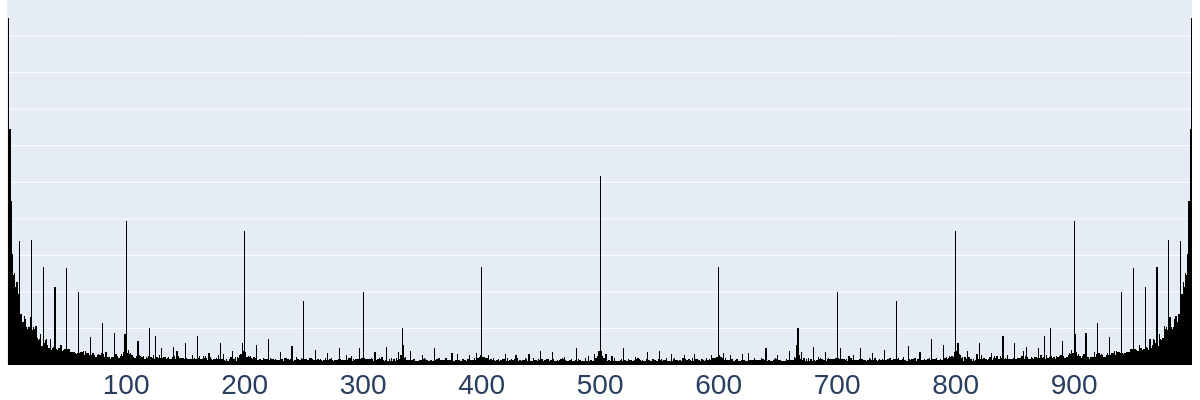}
        \caption{Llama 3 1B}
    \end{subfigure}
    \begin{subfigure}[h]{0.5\linewidth}
        \includegraphics[width=1.0\linewidth]{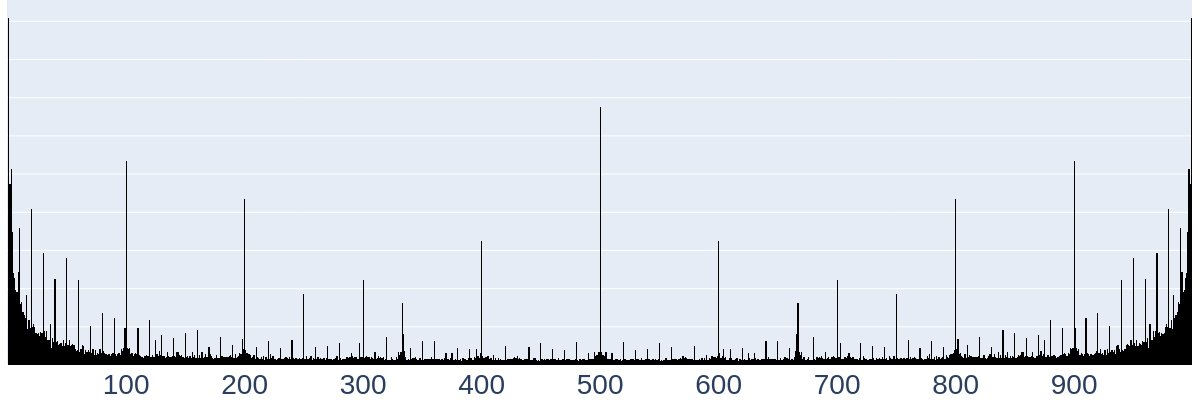}
        \caption{Llama 3 3B}
    \end{subfigure}
    \begin{subfigure}[h]{0.5\linewidth}
        \includegraphics[width=1.0\linewidth]{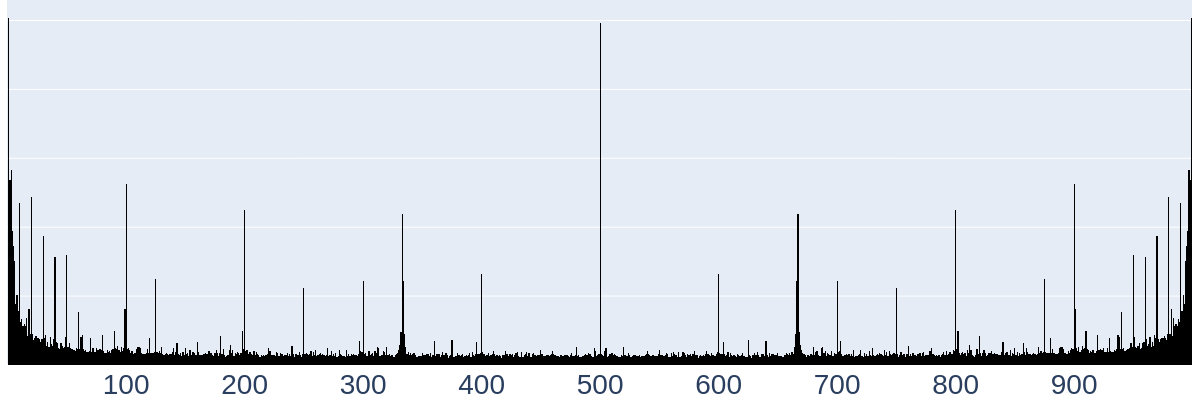}
        \caption{Llama 3 8B}
    \end{subfigure}
    \begin{subfigure}[h]{0.5\linewidth}
        \includegraphics[width=1.0\linewidth]{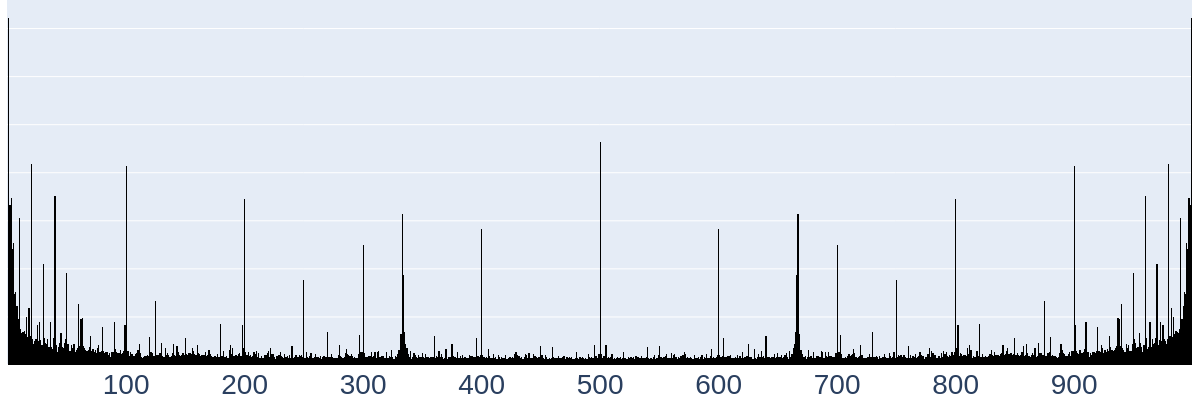}
        \caption{Llama 3 70B}
    \end{subfigure}
    \begin{subfigure}[h]{0.5\linewidth}
        \includegraphics[width=1.0\linewidth]{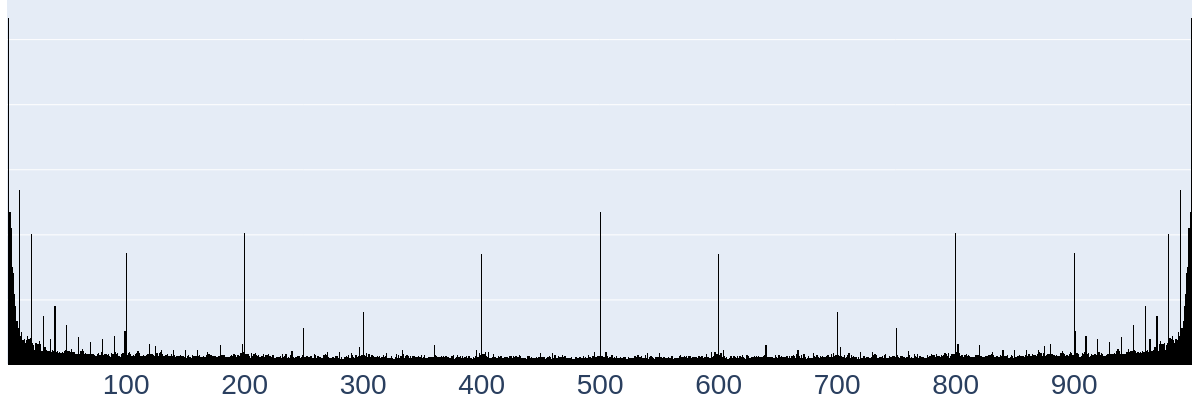}
        \caption{OLMo 2 1B}
    \end{subfigure}
    \begin{subfigure}[h]{0.5\linewidth}
        \includegraphics[width=1.0\linewidth]{figures/fig_embs_fft/embs_olmo1b_pca_fft.png}
        \caption{OLMo 2 7B}
    \end{subfigure}
    \begin{subfigure}[h]{0.5\linewidth}
        \includegraphics[width=1.0\linewidth]{figures/fig_embs_fft/embs_olmo1b_pca_fft.png}
        \caption{OLMo 2 13B}
    \end{subfigure}
    \begin{subfigure}[h]{0.5\linewidth}
        \includegraphics[width=1.0\linewidth]{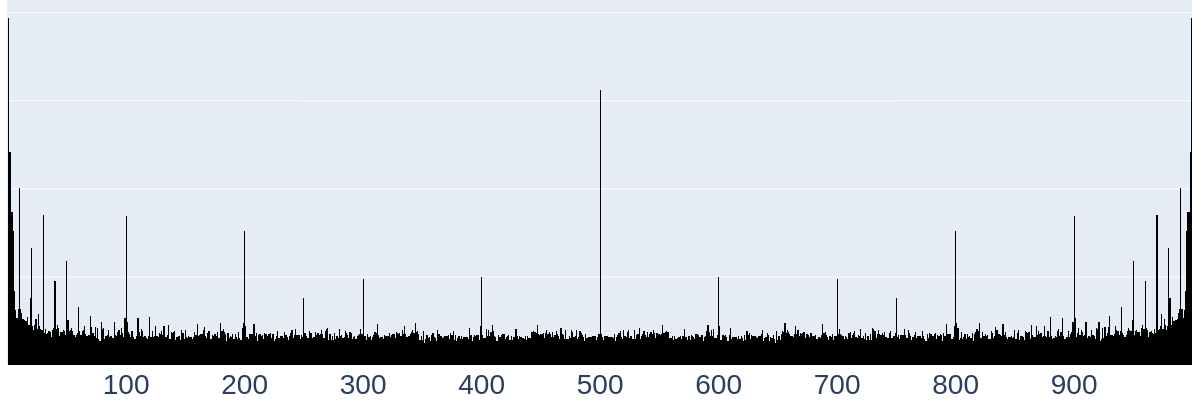}
        \caption{OLMo 2 32B}
    \end{subfigure}
    \begin{subfigure}[h]{0.5\linewidth}
        \includegraphics[width=1.0\linewidth]{figures/fig_embs_fft/embs_llama1b_pca_fft.png}
        \caption{Phi 4 15B}
    \end{subfigure}
    \caption{Maximal contribution (magnitude) of each Fourier base frequency's to embedding features in PCA (d=128) reduced space. Sparsity in this plot indicates strong regularity in the hidden structure of model embeddings. OLMo 2 32B has noticeably stronger contribution of all low-contribution frequencies, indicating high irregularity.}
    \label{fig:pca-fourier}
\end{figure*}

\subsection{Explainability plots for arithmetic tasks.}

\paragraph{Model behavior.}
To better explain the behavior of the LMs, we conduct a simple circuit analysis and a feature attribution experiment using integrated gradients \citep{pmlr-v70-sundararajan17a}.
For convenience, we focus on the two smaller models in our panel. OLMo 2 1B and Llama 3 1B. 

\begin{figure}
    \centering
    \begin{subfigure}[t]{0.475\linewidth}
        \centering
        \includegraphics[width=\textwidth]{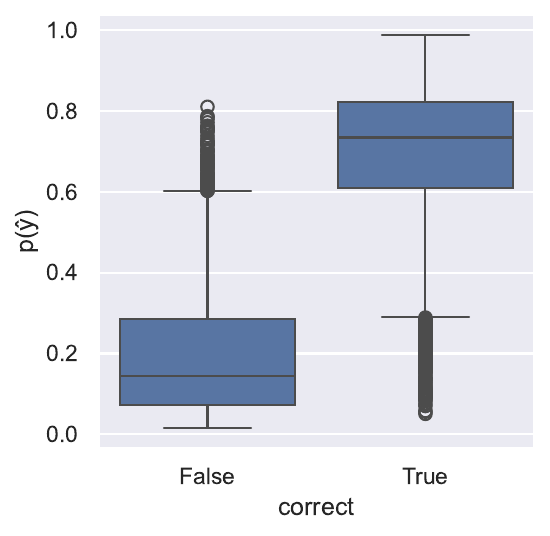}
        \caption{Llama 3 1B, addition.}
    \end{subfigure}%
    \quad
    \begin{subfigure}[t]{0.475\linewidth}
        \centering
        \includegraphics[width=\textwidth]{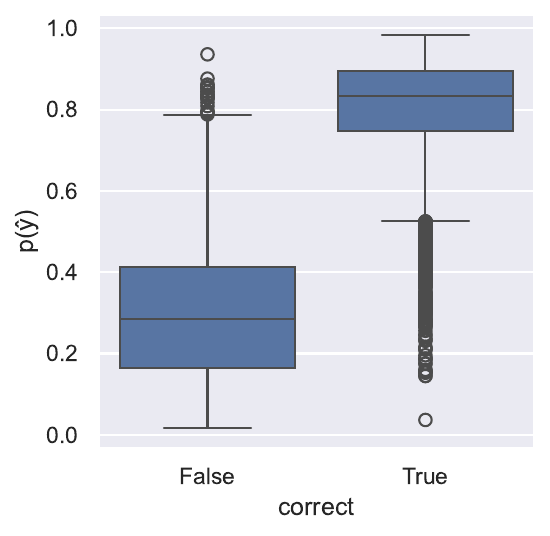}
        \caption{Llama 3 1B, subtraction.}
    \end{subfigure}
    
    \begin{subfigure}[t]{0.475\linewidth}
        \centering
        \includegraphics[width=\textwidth]{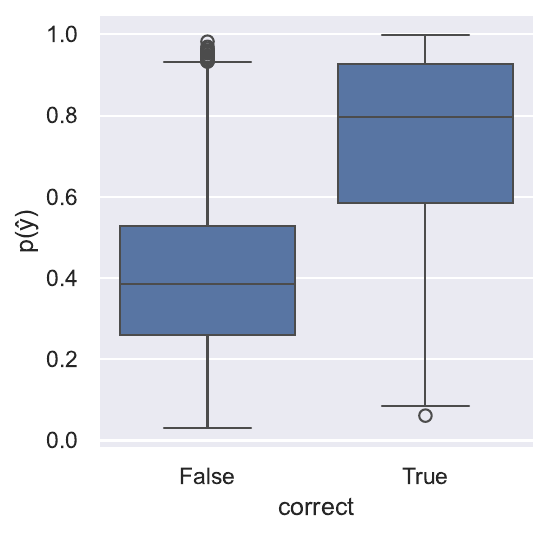}
        \caption{OLMo 2 1B, addition.}
    \end{subfigure}%
    \quad
    \begin{subfigure}[t]{0.475\linewidth}
        \centering
        \includegraphics[width=\textwidth]{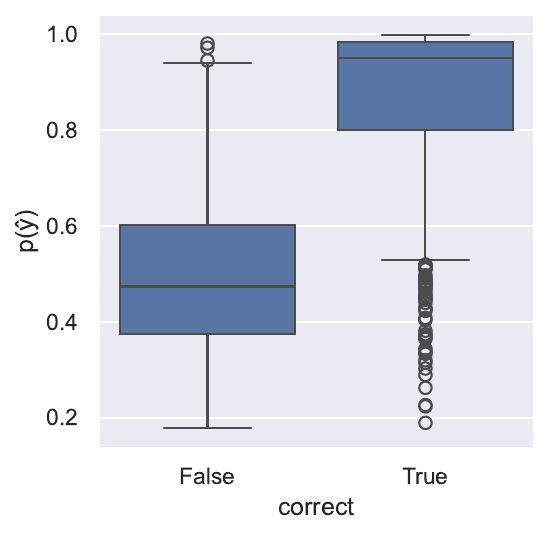}
        \caption{OLMo 2 1B, subtraction.}
    \end{subfigure}
    
    \caption{Probability mass on the predicted output token when the LLM yields a correct vs.~incorrect answer.}
    \label{fig:prob-mass}
\end{figure}

Both experiments suggest one major difference between operand pairs leading to failure and to success: the probability assigned by the LLM to the predicted output token tends to be statistically lower when the model produces an incorrect output, see \Cref{fig:prob-mass}.
We also observe the same subset of heads being activated for failure and success on the arithmetic task.
Besides the usefulness of this difference in probability mass for diagnostic purposes, these experiments also suggest a difference in \textit{degree} rather than \textit{kind} between failures and successes. 

\label{sec:explainability}

\begin{figure}
    \centering
    \begin{subfigure}[t]{\columnwidth}
        \centering
        \includegraphics[width=\columnwidth]{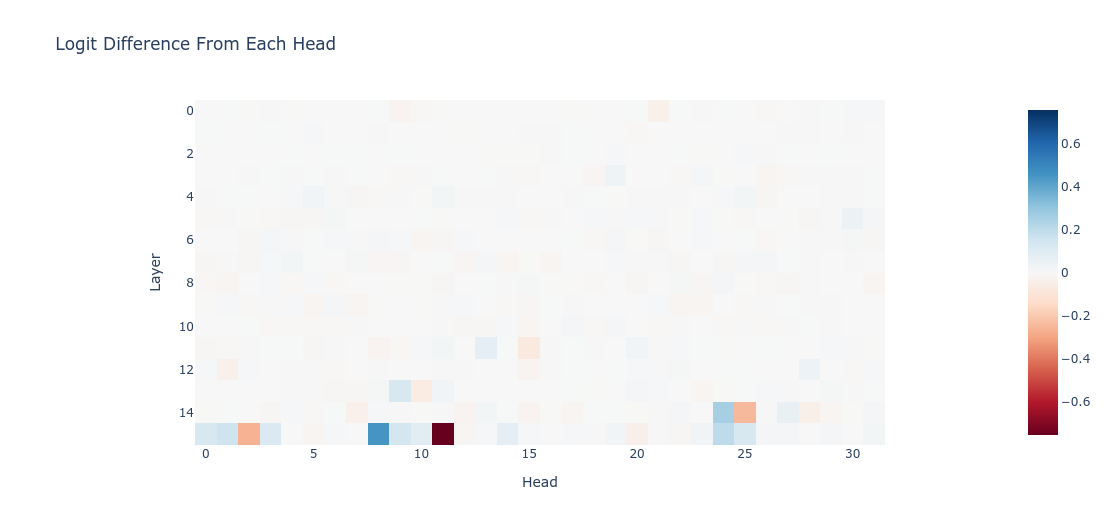}
        \caption{Llama 3 1B, addition performed correctly.}
    \end{subfigure}
    
    \begin{subfigure}[t]{\columnwidth}
        \centering
        \includegraphics[width=\columnwidth]{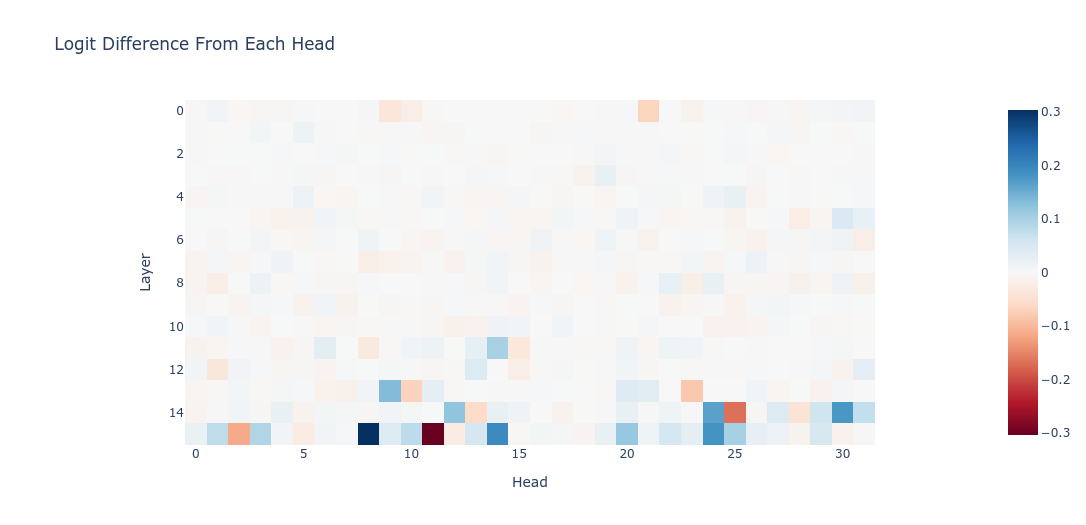}
        \caption{Llama 3 1B, addition performed incorrectly.}
    \end{subfigure}
    
    \begin{subfigure}[t]{\columnwidth}
        \centering
        \includegraphics[width=\columnwidth]{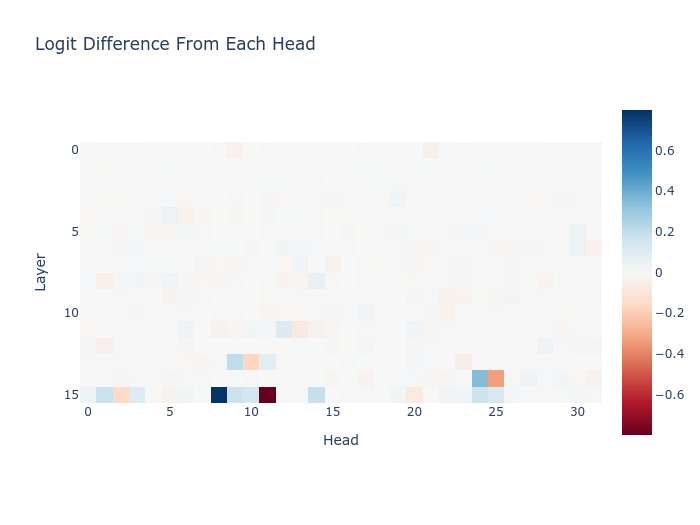}
        \caption{Llama 3 1B, subtraction performed correctly.}
    \end{subfigure}
    
    \begin{subfigure}[t]{\columnwidth}
        \centering
        \includegraphics[width=\columnwidth]{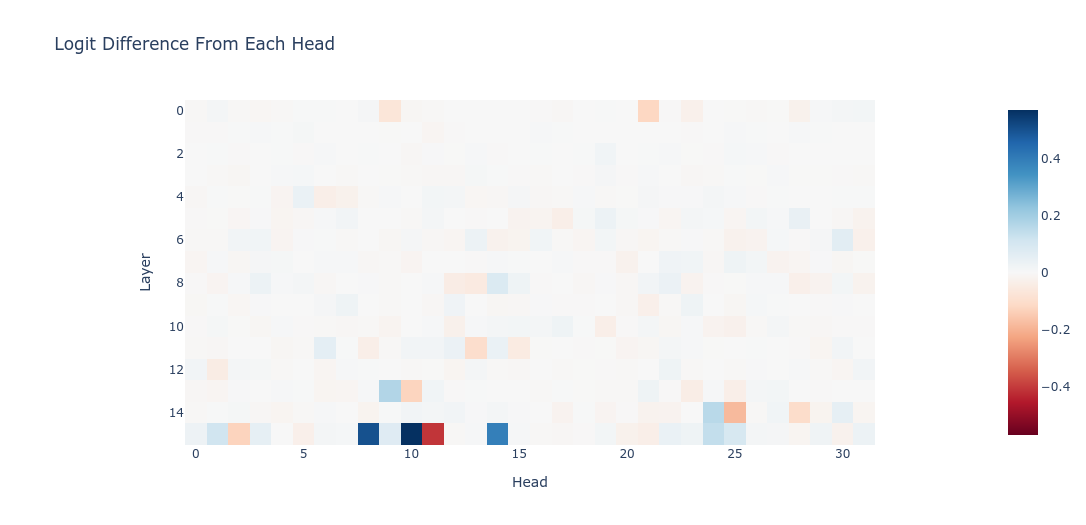}
        \caption{Llama 3 1B, subtraction performed incorrectly.}
    \end{subfigure}
    \caption{Head activations across arithmetic tasks for Llama 3 1B, broken down by task (addition and subtraction) and success (correct or incorrect computation.}
    \label{fig:active head}
\end{figure}

In \Cref{fig:active head}, we present an overview of head-level attribution of the logits in Llama 2 1B. The same heads in Layers 13 through 15 appear activated in all cases, playing the same inhibitor and booster roles. Incorrectly performed addition leads to a noisier overall pattern.
Remarkably, we observe that activity occurs in the latter stages of the model, whereas input embeddings (layer 0) already contain precise numeric information, as per our probing experiments.
This delayed processing may explain some of the errors we observe, despite the high accuracy of our probes in \Cref{sec:probes}.

\subsection{Analysis of sin-base probes' learned representations}
\label{appx:probe_weights}

\begin{figure*}[th]
    \centering
        \begin{subfigure}[t]{\textwidth}
            \includegraphics[width=\textwidth]{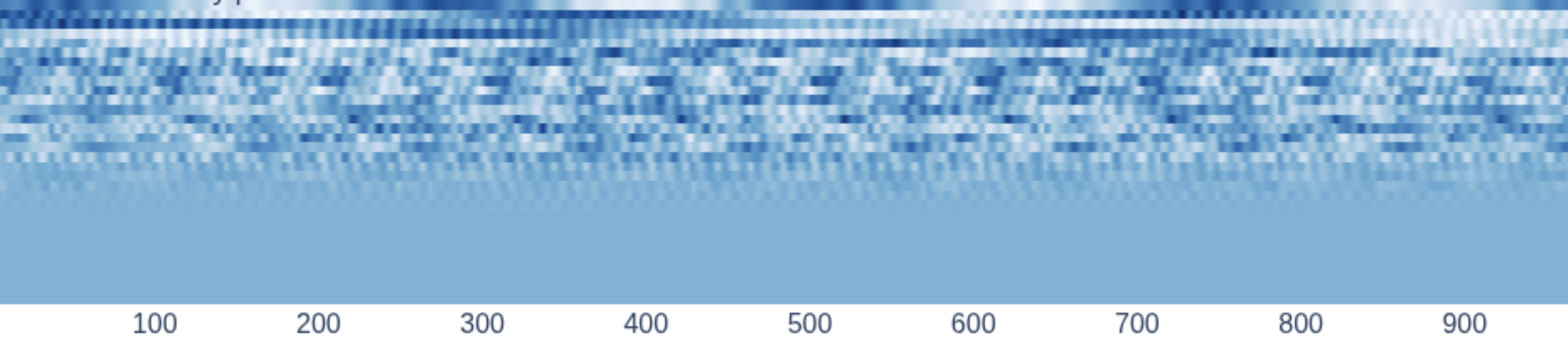}
            \caption{L1 regularization}
        \end{subfigure}
        
        \begin{subfigure}[t]{\textwidth}
            \centering
            \includegraphics[width=\textwidth]{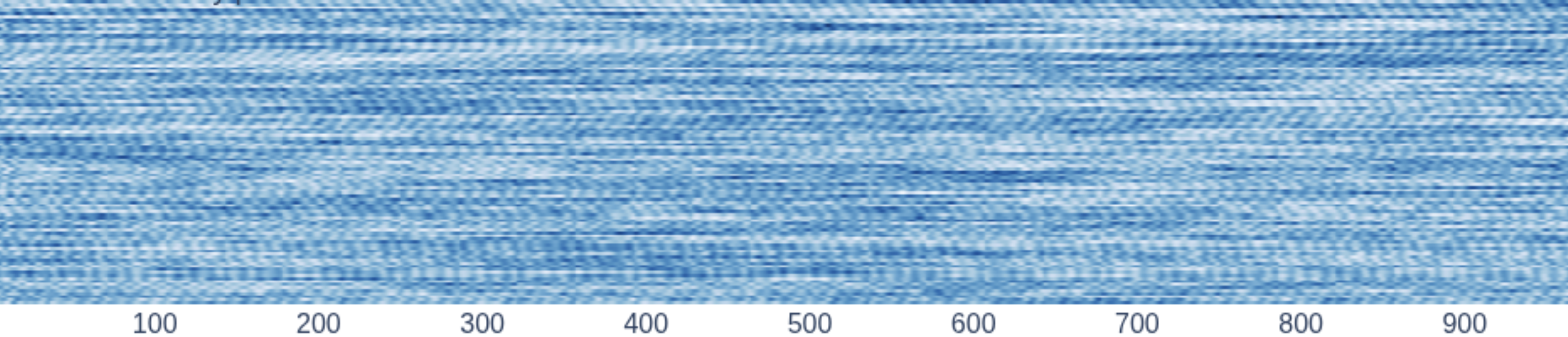}
            \caption{L2 regularization}
        \end{subfigure}
    
    \caption{Hidden representations of sin-base probes for numeric input embeddings of Llama 3 1B model, after training with different regularization strategies show that our sin-base probes learn to project numeric embeddings into a generalized, wave-like representation used as target inductive bias.}
    \label{fig:probe_weights}
\end{figure*}

In Figure~\ref{fig:probe_weights}, we can see that our newly proposed sin-like probes indeed learn to project input embeddings of models into an expected, generalized wave-like representation.

\section{Experimental details}
\label{sec:details}

\begin{table}[ht]
    \centering
    \begin{subfigure}[t]{\columnwidth}
        \centering
    \rowcolors{2}{white}{gray!50}
    \begin{tabular}{l l}
    \toprule
    1&``\texttt{$x_1$+$x_2$ equals to }'' \\
    2&``\texttt{The result of $x_1$+$x_2$ is }'' \\
    3&``\texttt{The result of $x_1$ plus $x_2$ is  }'' \\
    4&``\texttt{The result of $x_1$ plus $x_2$ = }''\\
    5&``\texttt{The result of $x_1$ plus $x_2$ =}''\\
    6&``\texttt{$x_1$ plus $x_2$ equals to }''\\
    7&``\texttt{$x_1$+$x_2$=}''\\
    8&``\texttt{$x_1$ plus $x_2$ equals }''\\
    9&``\texttt{$x_1$ plus $x_2$ is equal to }''\\
    10&``\texttt{$x_1$+$x_2$ equals }''\\
    11&``\texttt{$x_1$+$x_2$ is equal to }''\\
    12&``\texttt{$x_1$ plus $x_2$ equals }''\\
    13&``\texttt{$x_1$ plus $x_2$ is equal to }''\\
    \bottomrule
    \end{tabular}
    \caption{\label{tab:prompts addition}Prompts considered for addition task. $x_1$ and $x_2$ are placeholders for the augend and the addend. Prompts are delimited by double quotes; trailing white-space is significant.}
    \end{subfigure}
    
    \begin{subfigure}[t]{\columnwidth}
        \centering
    \rowcolors{2}{white}{gray!50}
    \begin{tabular}{l l}
    \toprule
    1&``\texttt{The result of $x_1$ minus $x_2$ is }'' \\
    2&``\texttt{The result of $x_1$ minus $x_2$ = }'' \\
    3&``\texttt{The result of $x_1$ minus $x_2$ =}'' \\
    4&``\texttt{$x_1$ minus $x_2$ equals to }'' \\
    5&``\texttt{$x_1$-$x_2$=}'' \\
    6&``\texttt{$x_1$ minus $x_2$ equals }'' \\
    7&``\texttt{$x_1$ minus $x_2$ is equal to }'' \\
    8&``\texttt{$x_1$-$x_2$ equals }'' \\
    9&``\texttt{$x_1$-$x_2$ is equal to }'' \\
    10&``\texttt{$x_1$ minus $x_2$ equals }'' \\
    11&``\texttt{$x_1$ minus $x_2$ is equal to }'' \\
    \bottomrule
    \end{tabular}
    \caption{
    \label{tab:prompts subtraction}Prompts considered for subtraction task. $x_1$ and $x_2$ are placeholders for the minuend and the subtrahend. Prompts are delimited by double quotes; trailing white-space is significant.}
    \end{subfigure}
    \caption{Prompts considered for engineering of arithmetic zero-shot setting.}
\end{table}

We conduct a minimal prompt optimization in \Cref{sec:add and subtract} to maximize the performances on arithmetic task. For all models below 20B parameters, we explore the prompts listed in \Cref{tab:prompts addition} and  \Cref{tab:prompts subtraction}, and report results with the highest performance in a zero-shot setting in \Cref{sec:add and subtract}. 

The most successful prompts for addition are prompt \#4 for Llama 3 1B, 2B and 8B as well as OLMo2 1B, and prompt \#3 for OLMO2 7B and 13B. As for subtraction, the most effective prompt was prompt \#1  for Llama 3 1B, OLMo2 1B and 7B, and prompt \#2 for Llama 3 3B and 8B as well as OLMo2 13B.

\section{Disclosure of usage of AI assistance}
We disclose that we used AI assistance during implementation of this work and its writing. Specifically, we used AI-based code auto-completion (Github Copilot) for increasing productivity of programming, and conversational chatbots (OpenAI ChatGPT, Google Gemini) for improving grammar and fluency of the text. We guarantee that all content is original and factually accurate.

\end{document}